\documentclass[runningheads,a4paper]{llncs}

\usepackage{url}
\usepackage{amssymb,amsmath}
\usepackage{graphicx}
\usepackage{subfigure} 
\usepackage{epstopdf}
\usepackage{algorithm}
\usepackage{algpseudocode}
\newtheorem{asm}{Assumption}

\urldef{\mailsa}\path|{ebrun, mazar}@cs.cmu.edu| 

\newcommand{\A}{\mathcal A}

\newcommand{\E}{\mathbb E}

\newcommand{\calS}{\mathcal{S}}

\newcommand{\hmu}{\widehat{\mu}}

\newcommand{\tmu}{\widetilde{\mu}}
\newcommand{\tpi}{\widetilde{\pi}}

\newcommand{\TODO}[1]{(\textbf{TODO: {#1}})}

\let\OldStatex\Statex
\renewcommand{\Statex}[1][3]{%
  \setlength\@tempdima{\algorithmicindent}%
  \OldStatex\hskip\dimexpr#1\@tempdima\relax}

\newcommand{\StatexIndent}[1][3]{%
  \setlength\@tempdima{\algorithmicindent}%
  \Statex\hskip\dimexpr#1\@tempdima\relax}
\algdef{S}[WHILE]{WhileNoDo}[1]{\algorithmicwhile\ #1}%

\begin{document}


\title{Regret Bounds for Reinforcement Learning with Policy Advice}

\titlerunning{Regret Bounds for Reinforcement Learning with Policy Advice}

%
%
\author{Mohammad Gheshlaghi Azar\inst{1} \and Alessandro Lazaric\inst{2} and Emma Brunskill\inst{1}}
%
\authorrunning{Azar, Lazaric, and Brunskill}

\institute{
Carnegie Mellon University, Pittsburgh, PA, USA
\\
\mailsa
\and INRIA Lille - Nord Europe, Team SequeL, Villeneuve d’Ascq, France
\\
\email{alessandro.lazaric@inria.fr}
}

%
%

\toctitle{Lecture Notes in Computer Science}
\tocauthor{Authors' Instructions}
\maketitle
\begin{abstract}
In some reinforcement learning problems an agent 
may be provided with a set of input policies, 
perhaps learned from prior experience or 
provided by advisors. We present a reinforcement 
learning with policy advice (RLPA) algorithm 
which leverages this input set and learns to 
use the best policy in the set for the 
reinforcement learning task at hand. We prove that 
RLPA has a sub-linear regret of $\widetilde O(\sqrt{T})$ relative to the best input 
policy, and that both this regret and its 
computational complexity are independent of 
the size of the state and action space.  
Our empirical simulations support our theoretical 
analysis. This suggests RLPA may offer 
significant advantages in large domains where 
some prior good policies are provided. 
\end{abstract}

\section{Introduction}
In reinforcement learning an agent seeks to learn a high-reward policy 
for selecting actions in a stochastic world without 
prior knowledge of the world dynamics model and/or reward 
function.  In this paper we consider 
when the agent is provided with an input set of potential policies, 
and the agent's objective is to perform as close as possible 
to the (unknown) best policy in the set. This scenario 
could arise when the general domain involves a finite set 
of types of RL tasks (such as different user models), 
each with known best policies, and the 
agent is now in one of the task types but doesn't know which one. 
Note that this situation could occur both in discrete state and 
action spaces, and in continuous state and/or action spaces: 
a robot may be traversing one of a finite set of different 
terrain types, but its sensors don't allow it to identify the 
terrain type prior to acting. Another example is when the 
agent is provided with a set of domain expert defined policies, 
such as stock market trading strategies. Since the agent 
has no prior information about which policy might perform 
best in its current environment, this remains a challenging 
RL problem. 

Prior research has considered the related case when 
an agent is provided with a fixed set of input (transition 
and reward) models, and the current domain is an (initially 
unknown) member of this 
set~\cite{Dyagilev2008EWRL,DiukICML2009,BrunskillAAMAS2012}. 
This actually provides the agent with more information than 
the scenario we consider (given a model we can extract a 
policy, but the reverse is not generally true), but more 
significantly, we find substantial theoretical and 
computational advantages from taking a model-free approach. 
Our work is also closely related to the idea of policy 
reuse~\cite{FernandezAAMAS2006}, where an agent tries to leverage 
prior policies it found for past tasks to improve performance 
on a new task; however, despite encouraging empirical 
performance, this work does not provide any formal guarantees. 
Most similar to our work is Talvitie and Singh's~\cite{talvitieIJCAI2007} 
AtEase algorithm which also learns to select among an input set of  
policies; however, in addition to algorithmic differences, 
 we provide a much more rigorous 
theoretical analysis that holds for a more general 
setting. 

We contribute a reinforcement learning with policy advice (RLPA) algorithm. 
RLPA is a model-free algorithm that, given an 
input set of policies, takes an optimism-under-uncertainty 
approach of adaptively selecting the policy that may have the 
highest reward for the current task. We prove the regret 
of our algorithm relative to the (unknown) best in the 
set policy scales with the square root of the time horizon, 
linearly with the size of the provided policy set, and 
is independent of the size of the state and 
action space. The computational complexity of our algorithm 
is also independent of the number of states and actions. 
This suggests our approach may have significant benefits 
in large domains over alternative approaches that typically scale with 
the size of the state and action space, and our 
preliminary simulation experiments provide empirical 
support of this impact.

\section{Preliminaries}

A Markov decision process (MDP) $M$ is defined as a tuple $\langle \calS, \A, P, r\rangle$ where $\calS$ is the set of states, $\A$ is the set of actions, $P:\calS\times\A\rightarrow \mathcal P(\calS)$ is the transition kernel mapping each state-action pair to a distribution over states, and $r:\calS\times\A \rightarrow \mathcal P([0,1])$ is the stochastic reward function mapping state-action pairs to a distribution over rewards bounded in the $[0,1]$ interval.\footnote{The 
extension to larger bounded regions $[0,d]$ is trivial and just 
introduces an additional $d$ multiplier to the resulting regret bounds.}
A policy $\pi$ is a mapping from states to actions. 
Two states $s_i$ and $s_j$ communicate with each other under policy $\pi$ if 
the probability of transitioning between $s_i$ and $s_j$ under $\pi$ is 
greater than zero. A state $s$ is recurrent under policy $\pi$ if the probability 
of reentering state $s$ under $\pi$ is 1. 
A recurrent  class is a set of 
recurrent states that all communicate with each other and no other states. Finally, a Markov process is unichain if its transition matrix consists of a single recurrent  class with (possibly) some transient states \cite[Chap. 8]{puterman1994markov}.

We define the performance of  
$\pi$ in a state $s$ as its expected average reward 
\begin{equation}\label{eq:avg.reward}
\mu^{\pi}(s)=\lim_{T\rightarrow\infty} \frac{1}{T} \mathbb E\bigg[\sum\nolimits_{t=1}^T r(s_t,\pi(s_t))\bigg|s_0=s\bigg],
\end{equation}
where $T$ is the number of time steps and the expectation is 
taken over the stochastic transitions and rewards. 
If $\pi$ induces a unichain Markov process on $M$, then $\mu^\pi(s)$ 
is constant over all the states $s\in\calS$, and we can 
define the bias function $\lambda^\pi$ such that
\begin{align}\label{eq:bias}
\lambda^\pi(s) + \mu^\pi = \E\big[r(s,\pi(s)) + \lambda^\pi(s')\big]. 
\end{align}
Its corresponding span is 
$sp(\lambda^\pi) = \max_{s} \lambda^\pi(s) - \min_{s} \lambda^\pi(s)$.  The bias $\lambda^\pi(s)$ can be seen as  the total difference between the reward of state $s$ and average reward.


In reinforcement learning~\cite{Sutton98} an agent does 
not know the transition $P$ and/or reward $r$ model in advance. Its 
goal is typically to find a policy $\pi$ that maximizes its obtained reward. 
In this paper, we consider reinforcement learning in an MDP $M$  
where the learning algorithm is 
provided with an input set of $m$ deterministic 
policies $\Pi=\{\pi_1,\ldots,\pi_m\}$. 
Such an input set of policies could arise 
in multiple situations, including: the policies may represent 
near-optimal policies for a set of $m$ 
MDPs $\{M_1,\ldots,M_m\}$ which may be related to the current MDP $M$; 
the policies may be the result of different approximation schemes (i.e., approximate policy iteration with different approximation spaces); or they may be provided by $m$ advisors. Our objective is to perform almost as well as 
the best policy in the input set $\Pi$  on the new 
MDP $M$ (with unknown $P$ and/or $r$). 

Our results require the following mild assumption: 
%
\begin{asm}\label{asm:best.policy}
There exists a policy $\pi^+\in\Pi$, which  induces a unichain Markov process on the MDP $M$, such that  the average reward   $\mu^+= \mu^{\pi^+} \geq \mu^\pi(s)$ for any state $s\in\calS$ and any policy $\pi\in\Pi$. 
We also assume that   $sp(\lambda^{\pi^+})\leq H$, where $H$ is a finite constant.\footnote{One can easily prove that the upper bound $H$ always exists  for any unichain Markov  reward process (see \cite[Chap. 8]{puterman1994markov}).}
\end{asm}
This assumption trivially holds when the optimal policy $\pi^*$ is in the set $\Pi$. Also, in those cases that all the policies in $\Pi$ induce some unichain Markov processes the existence of $\pi^+$ is guaranteed.\footnote{Note that Assumption \ref{asm:best.policy} in general is  a weaker assumption than assuming MDP $M$ is ergodic or unichain,
which would require that the induced Markov chains under \emph{all} policies be
recurrent or unichain, respectively: we only require that the best policy 
in the input set must induce a unichain Markov process.}


A popular measure of the performance of a reinforcement learning algorithm 
over $T$ steps is its regret relative to executing the 
optimal policy $\pi^*$ in $M$. 
We evaluate the regret relative to the best policy $\pi^+$ in the 
input set $\Pi$, 
\begin{align}\label{eq:regret}
\Delta(s) =  T\mu^{+}-\sum\nolimits_{t=1}^T r_t,
\end{align}
where $r_t\sim r(\cdot|s_{t},a_{t})$ and $s_0=s$. We notice that this definition of regret differs from the standard definition of regret by an (approximation) error $T(\mu^*-\mu^+)$ due to the possible sub-optimality of the policies in $\Pi$ 
relative to the optimal policy for MDP $M$. Further discussion on this definition is provided in Sec.~\ref{s:conclusions}. 







\section{Algorithm}\label{s:algorithm}
\begin{algorithm}[t!] 
\caption{Reinforcement Learning with Policy Advice (RLPA) }
\label{alg:B.MDP} 
\begin{algorithmic}[1]
\Require Set of policies $\Pi$, confidence $\delta$, span function $f$
\State Initialize $t=0$, $i=0$
\State Initialize $n(\pi)=1$, $\hmu(\pi)=0$, $R(\pi)=0$ and $K(\pi)=1$ for all $\pi\in\Pi$
\While {$t \leq T$}
\State Initialize $t_i = 0, \;T_i=2^i, \;\Pi_i=\Pi,\; \widehat H=f(T_i)$
\State $i=i+1$
\While { $t_i \leq T_i$ \& $\Pi_i\neq\emptyset$ }  \textbf{\textit{(run trial)}} 
\State $c(\pi) = (\widehat H+1) \sqrt{48\frac{\log (2t/\delta)}{n(\pi)}}+\widehat H \frac{K(\pi)}{n(\pi)}$
\State $B(\pi) = \hmu(\pi) + c(\pi)$
\State $\tpi=\arg\max_{\pi} B(\pi)$
\State $v(\widetilde\pi)=1$
\WhileNoDo{$t_i\leq T_i$ \& $v(\tpi)\!<\! n(\tpi)$ \&}
\State $\hmu(\tpi)-\frac{R(\widetilde \pi)}{n( \widetilde \pi)+v( \widetilde \pi)} \!\leq\!\! c(\tpi) \!+\! (\widehat H\!+\!1)\sqrt{48\frac {\log(2t/\delta)}{n( \widetilde \pi)+v(\widetilde \pi)}} + \widehat H \frac{K(\widetilde\pi)}{n(\widetilde\pi)+v(\widetilde \pi)}$ \algorithmicdo 
\State \textbf{\textit{(run episode)}}
\State $t=t+1$, $t_i = t_i + 1$
\State Take action $\tpi(s_t)$, observe $s_{t+1}$ and $r_{t+1}$ 
\State $v(\widetilde \pi)=v(\widetilde \pi)+1$ , $R(\widetilde \pi)= R(\widetilde \pi)+r_{t+1}$
\EndWhile
\State $K(\widetilde\pi)=K(\widetilde\pi)+1$
\If   {$\hmu(\tpi)-\frac{R(\widetilde \pi)}{n(\widetilde \pi)+v( \widetilde \pi)} >   c(\tpi) + (\widehat H\!+\!1)\sqrt{48\frac {\log(2t/\delta)}{n( \widetilde \pi)+v(\widetilde \pi)}}+\widehat H \frac{K(\widetilde\pi)}{n(\widetilde\pi)+v(\widetilde \pi)}$}
\State $\Pi_i=\Pi_i-\{\widetilde\pi\}$
\EndIf
\State $n(\widetilde \pi)=n(\widetilde \pi)+v(\widetilde \pi)$ , $\hmu(\tpi) = \frac{R(\tpi)}{n(\widetilde \pi)}$
\EndWhile
\EndWhile
\end{algorithmic} 
\end{algorithm}
In this section we introduce the Reinforcement Learning with Policy Advice (RLPA) algorithm (Alg.~\ref{alg:B.MDP}). 
Intuitively, the algorithm seeks to identify and use the policy in the 
input set $\Pi$ that yields the highest average reward on the 
current MDP $M$. As the average reward of each $\pi \in \Pi$ on 
$M$, $\mu^{\pi}$, is initially unknown, the algorithm proceeds by estimating 
these quantities by executing the different $\pi$ on the current 
MDP. More concretely, RLPA executes a series of trials, and within each trial is a series of episodes. 
Within each trial the algorithm selects the policies in $\Pi$ with the objective of effectively balancing between the exploration of all the policies 
in $\Pi$ and the exploitation of the most promising ones. 
Our procedure for doing this falls within the popular class 
of ``optimism in face  uncertainty'' methods. To do this, 
at the start of each episode, we 
define an upper bound on the possible average reward of 
each policy (Line 8): this average reward is computed as a 
combination of the average reward observed so far for this 
policy $\hat{\mu}(\pi)$, the number of time steps this 
policy has been executed $n(\pi)$ and $\widehat{H}$, which 
represents a guess of the span of the best policy, $H^+$. 
We then select the policy with the 
maximum upper bound $\tpi$ (Line 9) to run for this episode. 
Unlike in multi-armed bandit settings where a 
selected arm is pulled for only one step, here the 
MDP policy is run for up to $n(\pi)$ steps, 
i.e., until its total number of execution steps is at most doubled. 
If $\widehat{H} \geq H^+$ then the confidence 
bounds computed (Line 8) are valid confidence intervals 
for the true best policy $\pi^+$; however, they may 
fail to hold for any other policy $\pi$ whose span 
$sp(\lambda^\pi) \geq \widehat{H}$. Therefore, we 
can cut off execution of an episode when these 
confidence bounds fail to hold (the condition specified 
on Line 12), since the policy may not be an optimal 
one for the current MDP, if $\widehat{H} \geq H^+$.\footnote{See Sec. \ref{ss:gap.independent} for further discussion on the necessity of the condition on Line 12.}
In this case, we can eliminate the current policy 
$\tpi$ from the set of policies considered in this 
trial (see Line 20). After an episode terminates, 
the parameters of the current policy $\tpi$ 
(the number of steps $n(\pi)$ and average reward $\hmu(\pi)$) are updated,  
new upper bounds on the policies are computed, and the next 
episode proceeds. As the average reward estimates converge, 
the better policies will be chosen more. 

Note that since we do not know $H^+$ in advance, we 
must estimate it online: otherwise, if 
$\widehat{H}$ is not  a valid upper bound for the span $H^+$ (see Assumption~\ref{asm:best.policy}), a trial might eliminate 
the best policy $\pi^+$, thus incurring a significant regret. 
We address this by successively 
doubling the amount of time $T_i$ each trial is run, and defining 
a $\widehat{H}$ that is a function $f$ of the current trial length. See 
Sec.~\ref{ss:gap.independent} for a more detailed discussion on the choice of $f$. This procedure guarantees the algorithm will 
eventually find an upper bound on the span $H^+$
 and perform trials with very small regret in high probability. Finally, RLPA is an anytime algorithm since it does not need to know the time horizon $T$ in advance.

\section{Regret Analysis}\label{s:regret}
In this section we derive a regret analysis of RLPA and we compare its performance to existing RL regret minimization algorithms. We first derive preliminary results used in the proofs of the two main theorems.

We begin by proving a general high-probability bound on the difference between average reward $\mu^{\pi}$ and its empirical estimate $\hmu(\pi)$ 
of a policy $\pi$ (throughout this discussion we mean the average 
reward of a policy $\pi$ on a new MDP $M$). 
Let $K(\pi)$ be the number of episodes $\pi$ has been run, each of them of length $v_k(\pi)$ ($k=1,\ldots,K(\pi)$). The empirical average $\hmu(\pi)$ is defined as
\begin{align}\label{eq:empirical.average}
\hmu(\pi) = \frac{1}{n(\pi)} \sum\nolimits_{k=1}^{K(\pi)} \sum\nolimits_{t=1}^{v_k(\pi)} r_t^k,
\end{align}
where $r_t^k \sim r(\cdot | s_t^k, \pi(s_t^k))$ is a random sample of the reward observed by taking the action suggested by $\pi$ and $n(\pi) = \sum_k v_k(\pi)$ is the total count of samples. Notice that in each episode $k$, the first state $s_1^k$ does not necessarily correspond to the next state of the last step $v_{k-1}(\pi)$ of the previous episode. 
\begin{lemma} \label{lem:conc:Markov}   
Assume that a policy $\pi$ induces on the MDP $M$ a single recurrent class with some additional transient states, i.e., $\mu^{\pi}(s)=\mu^{\pi}$ for all $s\in\calS$. Then the difference between the average reward and its empirical estimate (Eq.~\ref{eq:empirical.average}) is
\begin{equation*}
|\hmu(\pi)- \mu^{\pi}| \leq 2(H^{\pi}+1)\sqrt{ \dfrac{2\log(2/\delta)}{n(\pi)}} + H^\pi \dfrac{K(\pi)}{n(\pi)},
\end{equation*}
with probability $\geq 1-\delta$, where $H^\pi = sp(\lambda^\pi)$ (see Eq.~\ref{eq:bias}).
\end{lemma}
\begin{small}
\begin{proof}
Let $r_{\pi}(s_t^k)= \mathbb E (r_t^k|s_t^k,\pi(s_t^k))$, $\epsilon_r(t,k)=r_t^k-r_{\pi}(s_t^k)$, and $P^{\pi}$ be the state-transition kernel under policy $\pi$ (i.e. for finite state and action spaces, $P^{\pi}$ is the $|S| \times |S|$ matrix where the $ij$-th entry is $p(s_j|s_i,\pi(s_i))$). 
Then we have
\begin{align*}
\hmu(\pi)- \mu^{\pi}&= \frac{1}{n(\pi)} \bigg(\sum_{k=1}^{K(\pi)}\sum_{t=1}^{v_k(\pi)} (r_t^k-\mu^{\pi}) \bigg) =\frac{1}{n(\pi)} \bigg(\sum_{k=1}^{K(\pi)}\sum_{t=1}^{v_k(\pi)} (\epsilon_r(t,k)+r_{\pi}(s_t^k)-\mu^{\pi})\bigg) \\
&=\frac{1}{n(\pi)} \bigg( \sum_{k=1}^{K(\pi)}\sum_{t=1}^{v_k(\pi)} (\epsilon_r(t,k)+\lambda^{\pi}(s_t^k)-P^{\pi}\lambda^{\pi}(s_t^k)) \bigg),
\end{align*}
where the second line follows from Eq.~\ref{eq:bias}.
Let $\epsilon_{\lambda}(t,k)=\lambda^{\pi}(s_{t+1}^k)-P^{\pi}\lambda^{\pi}(s_t^k)$. Then we have
\begin{align*}
\hmu(\pi)- \mu^{\pi}&= \frac{1}{n(\pi)} \bigg( \sum_{k=1}^{K(\pi)}\sum_{t=1}^{v_k(\pi)} (\epsilon_r(t,k)+\lambda^\pi(s_{t+1}^k) - \lambda^\pi(s_{t+1}^k) + \lambda^{\pi}(s_t^k)-P^{\pi}\lambda^{\pi}(s_t^k)) \bigg) \\
&\leq\frac{1}{n(\pi)} \bigg(\sum_{k=1}^{K(\pi)}(H^{\pi}+\sum_{t=1}^{v_k(\pi)} \epsilon_r(t,k)+\sum_{t=1}^{v_k(\pi)-1}\epsilon_{\lambda}(t,k)) \bigg),
\end{align*}
where we bounded the telescoping sequence 
$\sum_t (\lambda^{\pi}_{s_t^k} - \lambda^\pi(s_{t+1}^k))  \leq sp(\lambda^\pi) = H^\pi$.
The sequences of random variables $\{\epsilon_r\}$ and $\{\epsilon_{\lambda}\}$, as well as their sums,  are martingale difference sequences. 
Therefore we can apply Azuma's inequality and obtain the bound
\begin{align*}
\hmu(\pi)- \mu^{\pi}&\leq\dfrac{ K(\pi) H^{\pi}+2\sqrt{2 n(\pi) \log(1/\delta)} + 2H^{\pi}\sqrt{ 2(n(\pi)-K(\pi))\log(1/\delta)}}{n(\pi)}
\\
&\leq H^\pi \dfrac{K(\pi)}{n(\pi)}+2(H^{\pi}+1)\sqrt{ \dfrac{2\log(1/\delta)}{n(\pi)}},
\end{align*}
with probability $\geq 1-\delta$, where in the first inequality we bounded the error terms $\epsilon_r$, each of which is bounded in $[-1,1]$, and $\epsilon_\lambda$, bounded in $[-H^\pi,H^\pi]$. The other side of the inequality follows exactly the same steps. \qed
\end{proof}
\end{small}
In the algorithm $H^\pi$ is not known and at each trial $i$ the confidence bounds are built using the guess on the span $\widehat{H} = f(T_i)$, where $f$ is an increasing function. 
For the algorithm to perform well, it needs to not 
discard the best policy $\pi^+$ (Line 20). The following lemma guarantees that after a certain number of steps, with high probability the policy $\pi^+$ is not discarded in any trial.
\begin{lemma}\label{lem:inc.pi}
For any trial started after $T \geq T^+ = f^{-1}(H^{+})$, the probability of policy $\pi^+$ to be excluded from $\Pi_A$ at anytime is less than $(\delta/T)^{6}$.
\end{lemma}
\begin{small}
\begin{proof}
Let $i$ be the first trial such that $T_i \geq f^{-1}(H^{+})$, which implies that $\widehat{H}=f(T_i) \geq H^+$. The corresponding step $T$ is at most the sum of the length of all the trials before $i$, i.e., $T \leq \sum_{j=1}^{i-1} 2^j \leq 2^i$, thus leading to the condition $T \geq T^+ = f^{-1}(H^{+})$. 
After $T \geq T^+$ the conditions in Lem.~\ref{lem:conc:Markov} (with Assumption~\ref{asm:best.policy}) are satisfied for $\pi^+$. Therefore  
the confidence intervals hold with probability 
at least $1-\delta$ and we have for $\hmu(\pi^+)$
\begin{align*}
\hmu(\pi^+)- \mu^+ &\leq 2(H^{+} +1)\sqrt{ \dfrac{2\log(1/\delta)}{n(\pi^+)}} + H^+ \dfrac{K(\pi^+)}{n(\pi^+)}
 \\&\leq 2(\widehat{H}+1)\sqrt{ \dfrac{2\log(1/\delta)}{n(\pi^+)}} + \widehat{H} \dfrac{K(\pi^+)}{n(\pi^+)},
\end{align*}
where $n(\pi^+)$ is number of steps when policy $\pi^+$ has been selected until $T$. Using a similar argument as in the proof of 
Lem.~\ref{lem:conc:Markov}, we can derive 
\begin{equation*}
\mu^+-\frac{R(\pi^+)}{n(\pi^+) + v(\pi^+)} \leq 2(\widehat{H}+1)\sqrt{ \dfrac{2\log(1/\delta)}{n(\pi^+) + v(\pi^+)}} + \widehat H \frac{K(\pi^+)}{n(\pi^+)+v(\pi^+)}, 
\end{equation*}
with probability at least $1-\delta$. 
Bringing together these two conditions, and applying the union bound,  
we have that the condition on Line12 holds with at least probability 
$1-2 \delta$ and thus $\pi^+$ is never discarded. 
More precisely Algo.~\ref{alg:B.MDP} uses slightly larger confidence intervals (notably $\sqrt{48\log(2t/\delta)}$ instead of $2\sqrt{2\log(1/\delta)}$), which guarantees that $\pi^+$ is discarded with at most a probability of $(\delta/T)^{6}$. \qed
\end{proof}
\end{small}
We also need the $B$-values (line 9) to be valid upper confidence bounds on the average reward of the best policy $\mu^+$.
\begin{lemma}\label{lem:optimism}
For any trial started after $T \geq T^+ = f^{-1}(H^+)$, the $B$-value of $\tpi$ is an upper bound on $\mu^+$
with probability $\geq 1-(\delta/T)^6$.
\end{lemma}
\begin{small}
\begin{proof}
Lem.~\ref{lem:inc.pi} guarantees that the policy $\pi^+$ is in $\Pi_A$ w.p. $(\delta/T)^{6}$. This combined with Lem.~\ref{lem:conc:Markov} and the fact that $f(T)>H^+$ implies that the $B$-value $B(\pi^+) = \hmu(\pi^+)+c(\pi^+)$ is a high-probability upper bound on $\mu^+$ and, since $\tpi$ is the policy with the maximum $B$-value, the result follows. \qed
\end{proof}
\end{small}
Finally, we bound the total number of episodes a policy could be selected.
\begin{lemma} \label{lem:episodes}   
After $T \geq T^+ = f^{-1}(H^+)$ steps of Algo.~\ref{alg:B.MDP}, 
let $K(\pi)$ be the total number of episodes $\pi$ has been selected and $n(\pi)$ the corresponding total number of samples, then
\begin{equation*}
K(\pi) \leq \log_2( f^{-1}(H^+))+\log_2(T)+\log_2(n(\pi)),
\end{equation*}
with probability $\geq 1-(\delta/T)^6$.
\end{lemma}
\begin{small}
\begin{proof}
Let $n_k(\pi)$ be the total number of samples at the beginning of episode $k$ (i.e., $n_k(\pi) = \sum_{k'=1}^{k-1} v_{k'}(\pi)$). In each trial of Algo.~\ref{alg:B.MDP}, an episode is terminated 
when the number of samples is doubled (i.e., $n_{k+1}(\pi) = 2n_k(\pi)$), or when the consistency condition (last condition on Line12) is violated and the policy is discarded or the trial is terminated (i.e., $n_{k+1} \geq n_k(\pi)$). We denote by $\overline{K}(\pi)$ the total number of episodes truncated before the number of samples is doubled, then $n(\pi)\geq 2^{K(\pi)-\overline{K}(\pi)}$. Since the episode is terminated before the number of samples is doubled only when either the trial terminates or the policy is discarded, in each trial this can only happen once per policy. Thus we can bound $\overline{K}(\pi)$ by the number of trials. A trial can either terminate because its maximum length $T_i$ is reached or when all the polices are discarded (line 6). From Lem.~\ref{lem:inc.pi}, we have that after $T \geq f^{-1}(H^+)$, $\pi^+$ is never discarded w.h.p. and a trial only terminates when $t_i>T_i$. Since $T_i = 2^i$, it follows that the number of trials is bounded by $\overline{K}(\pi) \leq \log_2({f}^{-1}(H^+))+\log_2(T)$. So, we have $n(\pi)\geq 2^{K(\pi)- \log_2({f}^{-1}(H^+))-\log_2(T)}$, which implies the statement of the lemma. \qed
\end{proof}
\end{small}
Notice that if we plug this result in the statement of Lem.~\ref{lem:conc:Markov}, we have that the second term converges to zero faster than the 
first term which decreases as $O(1/\sqrt{n(\pi)})$, thus in principle it could be possible to use alternative episode stopping criteria, such as $v(\pi) \leq \sqrt{n(\pi)}$. But while this would not significantly affect the convergence rate of $\hmu(\pi)$, it may worsen the global regret performance in Thm.~\ref{thm:gap.independent}.
%
%
\subsection{Gap-Independent Bound}\label{ss:gap.independent}
We are now ready to derive the first regret bound for RLPA.
\begin{theorem}\label{thm:gap.independent}
Under Assumption~\ref{asm:best.policy} for any $T \geq T^+=f^{-1}(H^+)$ the regret of Algo.~\ref{alg:B.MDP} is bounded as
\begin{align*}
\Delta(s) \leq 24(f(T)+1)\sqrt{3Tm(\log(T/\delta))}+\sqrt{T}+6f(T)m(\log_2(T^+)+2\log_2(T)),
\end{align*}
%
with probability at least $1-\delta$ for any initial state $s\in\calS$.
\end{theorem}
\begin{small}
\begin{proof}
We begin by bounding the regret from executing each policy $\pi$. We consider the $k(\pi)$-th episode when policy $\pi$ has been selected (i.e., $\pi$ is the optimistic policy $\tpi$) and we study its corresponding 
total regret $\Delta_{\pi}$. We denote by $n_k(\pi)$ the number of steps of policy $\pi$ at the beginning of  episode $k$ and 
$v_k(\pi)$ the number of steps in episode $k$. 
Also at time step $T$, 
let the total number of episodes,   
$v_k(\pi)$ and $n_k$, for each policy $\pi$ be denoted as 
$K(\pi)$, $v(\pi)$  and $n(\pi)$ respectively.  
We also let $\pi\in\Pi$, $B(\pi)$, $c(\pi)$, $R(\pi)$ and $\hmu(\pi)$ 
be the latest values of these variables at time step $T$ 
for each policy $\pi$. Let $\mathcal E=\{\forall t=f^{-1}(H^+),\dots,T,\pi^+ \in \Pi_{A} \enspace \& \enspace \tpi \geq \mu^+ \}$ be the event 
under which $\pi^+$ is never removed from the set of 
policies $\Pi_A$, and where the upper bound of the 
optimistic policy $\tpi$, $B(\tpi)$, is always as large 
as the true average reward of the best policy $\mu^+$. 
On the event $\mathcal E$, $\Delta_{\pi}$ can be bounded as 
\begin{align*}
\Delta_{\pi}&=  \sum_{k=1}^{ K(\pi)}\sum_{t=1}^{ v_k( \pi)} (\mu^+-r_t ) \overset{(1)}{\leq}  \sum_{k=1}^{ K(\pi)}\sum_{t=1}^{ v_k( \pi)} ( B(\pi)- r_t )\leq (n(\pi)+v(\pi)) (\hmu(\pi)+c(\pi)) - R(\pi)
\\&\overset{(2)}{\leq}  (n(\pi)+v(\pi))\left( 3(f(T)+1)\sqrt{48\frac {\log(T/\delta)}{n(\pi)}}+3f(T) \frac{K(\pi)}{n(\pi)}\right) 
 \\
&\overset{(3)}{\leq} 24(f(T)+1)\sqrt{3 n( \pi)\log (T/\delta)} +6f(T)K(\pi),
\end{align*}
where in (1) we rely on the fact that $\pi$ is only executed when 
it is the optimistic policy, and $B(\pi)$ is optimistic 
with respect to $\mu^+$ according to Lem.~\ref{lem:optimism}. 
(2) immediately follows from the stopping condition at Line 12 and the definition of $c(\pi)$.  (3) follows from the condition on doubling the samples (Line 12) which guarantees $v(\pi) \leq n( \pi)$.  

We now bound the total regret $\Delta$ by summing over all the policies.
\begin{align*}
\Delta &=  \sum_{\pi\in\Pi}24(f(T)+1)\sqrt{3n( \pi)\log(T/\delta)} +6 f(T) \sum_{\pi\in\Pi} K(\pi)
\\
&\overset{(1)}{\leq}
 24 (f(T)+1) \sqrt{3m \sum_{\pi\in\Pi}n( \pi)\log(T/\delta)} +6f(T) \sum_{\pi\in\Pi} K(\pi)
\\& \overset{(2)}{\leq}
24 (f(T)+1) \sqrt{3m T \log (T/\delta)} +6f(T) m (\log_2( f^{-1}(H^+))+2\log_2(T)),
\end{align*}
where in $(1)$ we use Cauchy-Schwarz inequality and (2) follows from $\sum_{\pi}n(\pi)\leq T$, Lem.~\ref{lem:episodes}, and 
 $\log_2(n(\pi)) \leq \log_2(T)$.

Since $T$ is an unknown time horizon, we need to provide a bound which holds with high probability uniformly over all the possible values of $T$. Thus we need to deal with the case when $\mathcal E$ does not hold. Based  on Lem. 1 and by following  similar lines to~\cite{UCRLAuer}, we can prove that the total regret of the episodes in which the true model is discarded is bounded by $\sqrt{T}$ with probability  at least $1-\delta/(12T^{5/4})$. 
Due to space limitations, we omit the details, but we can then 
prove the final result by 
combining the regret in both cases (when $\mathcal E$ holds or 
does not hold) 
and taking union bound on all possible values of $T$.   \qed
\end{proof}
\end{small}
%
%
%
%
A significant advantage of RLPA over generic RL algorithms 
(such as UCRL2) is that the regret of RLPA is independent of 
the size of the state and action spaces: in contrast, 
the regret of UCRL2 scales as $O(S\sqrt{AT})$. 
This advantage is obtained by exploiting the prior information that $\Pi$ contains good policies, which allows the algorithm to focus on testing their performance to identify the best, instead of building an estimate of the current MDP over the whole state-action space as in UCRL2. It is also 
informative to compare this result to other methods using some form of prior knowledge. In~\cite{maillard2013optimal} the objective is to learn the optimal policy along with a state representation which satisfies the Markov property. The algorithm receives as input a set of possible state representation models and under the assumption that one of them is Markovian, the algorithm is shown to have a sub-linear regret. Nonetheless, the algorithm inherits the regret of UCRL itself and still displays a $O(S\sqrt{A})$ dependency on states and actions. In~\cite{Dyagilev2008EWRL} the Parameter Elimination (PEL) algorithm is provided with a set of MDPs. The algorithm is analyzed in the PAC-MDP framework and under the assumption that the true model actually belongs to the set of MDPs, it is shown to have a performance which does not depend on the size of the state-action space and it only has a $O(\sqrt{m})$ a dependency on the number of MDPs $m$.\footnote{Notice that PAC bounds are always squared w.r.t. regret bounds, thus the original $m$ dependency in~\cite{Dyagilev2008EWRL} becomes $O(\sqrt{m})$ when compared to a regret bound.} In our setting, although no model is provided and no assumption on the optimality of $\pi^*$ is made, RLPA achieves the same dependency on $m$.

The span $sp(\lambda^\pi)$ of a policy is known to be a critical parameter determining how well and fast the average reward of a policy can be estimated using samples (see e.g.,~\cite{bartlett2009regal}). In Thm.~\ref{thm:gap.independent} we show that only the span $H^+$ of the best policy $\pi^+$ affects the performance of RLPA even when other policies have much larger spans. Although this result may seem surprising (the algorithm estimates the average reward for all the policies), it follows from the use of the third condition on Line12 where an episode is terminated, and a policy is discarded, whenever the empirical estimates are not consistent with the guessed confidence interval. Let us consider the case when $\widehat{H} > H^+$ but $\widehat{H} < sp(\lambda^\pi)$ for a policy which is selected as the optimistic policy $\tpi$. Since the confidence intervals built for $\pi$ are not correct (see Lem.~\ref{lem:conc:Markov}), $\tpi$ could be selected for a long while before selecting a different policy. On the other hand, the condition on the consistency of the observed rewards would discard $\pi$ (with high probability), thus increasing the chances of the best policy (whose confidence intervals are correct) to be selected. We also note that 
 $H^+$ appears as a constant in the regret through $\log_2(f^{-1}(H^+))$ and this suggests that the optimal choice of $f$ is $f(T) = \log(T)$, which would lead to a bound of order (up to constants and logarithmic terms) $\widetilde O(\sqrt{Tm} + m)$.

\subsection{Gap-Dependent Bound}
Similar to~\cite{UCRLAuer}, we can derive an alternative bound for RLPA where the dependency on $T$ becomes logarithmic and the gap between the average of the best and second best policies appears. We first need to introduce two assumptions.
\begin{asm}[Average Reward]
\label{asm:avg.reward}
Each policy $\pi\in\Pi$ induces on the MDP $M$ a single recurrent class with some additional transient states, i.e., $\mu^{\pi}(s)=\mu^{\pi}$ for all $s\in\calS$. This implies that $H^\pi = sp(\lambda^\pi) < +\infty$.
\end{asm}
\begin{asm}[Minimum Gap]
\label{asm:gap}
Define the gap between the average reward of the best policy $\pi^+$ and the average reward of any other policy as $\Gamma(\pi,s)=\mu^{+}-\mu^{\pi}(s)$ for all $s\in\calS$. We then assume that for all $\pi\in \Pi-\{\pi^{+}\}$ and $s\in\calS$, $\Gamma(\pi,s)$ is uniformly bounded from below by a positive constant $\Gamma_{\min}>0$.
\end{asm}
%
%
\begin{theorem}[Gap Dependent  Bounds]
\label{thm:gap.dependent}
Let Assumptions \ref{asm:avg.reward} and \ref{asm:gap} hold.  Run  Algo.~\ref{alg:B.MDP} with the choice of $\delta=\sqrt[3]{ 1/T}$ (the stopping time $T$ is  assumed to be known here).  Assume that for all $\pi\in \Pi$ we have that  $H_{\pi}\leq H_{\max}$. Then the expected regret of Algo.~\ref{alg:B.MDP}, after $T\geq T^+ = f^{-1}(H^{+})$ steps,  is bounded as
\begin{align}\label{eq:gap.dependent}
\mathbb E(\Delta(s))= O\bigg( m \frac{(f(T)+H_{\max})(\log_2(mT)+\log_2(T^+))}{\Gamma_{\min}} \bigg),
\end{align}
for any initial state $s\in\calS$.
\end{theorem}

\begin{small}
\begin{proof} \textbf{(sketch)}
Unlike for the proof of Thm.~\ref{thm:gap.independent}, here we need a more refined control on the number of steps of each policy as a function of the gaps $\Gamma(\pi,s)$. We first notice that Assumption~\ref{asm:avg.reward} allows us to define $\Gamma(\pi) = \Gamma(\pi,s) = \mu^+ - \mu^\pi$ for any state $s\in\calS$ and any policy $\pi\in\Pi$. We consider the high-probability event $\mathcal E=\{\forall t=f^{-1}(H^+),\dots,T,\pi^{+} \in \Pi_{A}\}$ (see Lem.~\ref{lem:inc.pi}) where for all the trials run after $f^{-1}(H^+)$ steps never discard policy $\pi^+$. We focus on the episode at time $t$, when an optimistic policy $\tpi\neq\pi^+$ is selected for the $k(\pi)$-th time, and we denote by $n_k(\tpi)$ the number of steps of $\tpi$ before episode $k$ and $v_k(\pi)$ the number of steps during episode $k(\pi)$. The cumulative reward during episode $k$ is $R_k(\tpi)$ obtained as the sum of $\hmu_k(\tpi)n_k(\tpi)$ (the previous cumulative reward) and the sum of $v_k(\tpi)$ rewards received since the beginning of the episode.  Let $\mathcal E=\{\forall t=f^{-1}(H^+),\dots,T,\pi^+ \in \Pi_{A} \enspace \& \enspace \tpi \geq \mu^+ \}$ be the event 
under which $\pi^+$ is never removed from the set of 
policies $\Pi_A$, and where the upper bound of the 
optimistic policy $\tmu$, $B(\tpi)$, is always as large 
as the true average reward of the best policy $\mu^+$. On event $\mathcal E$  we have

\begin{align*}
&3(\widehat{H}+1)\sqrt{48\frac {\log(t/\delta)}{n_k(\tpi)}}+3\frac{k(\pi)}{n_k(\tpi)} \overset{(1)}{\geq} B(\tpi)-\frac{R_k(\tpi)}{n_k(\tpi)+v_k(\tpi)} \\
\overset{(2)}{\geq} &\mu^+ - \frac{R_k(\tpi)}{n_k(\tpi)+v_k(\tpi)}\geq \mu^+ - \mu^{\tpi} + \frac{1}{n_k(\tpi)+v_k(\tpi)} \sum_{t=1}^{n_k(\tpi)+v_k(\tpi)}(\mu^{\tpi}-r_{t})
\\
\overset{(3)}{\geq}&\Gamma_{\min}+\frac{1}{n_k(\tpi)+v_k(\tpi)} \sum_{t=1}^{n_k(\tpi)+v_k(\tpi)}(\mu^{\tpi}-r_{t})\overset{(4)}{\geq}\Gamma_{\min}- H^{\widetilde \pi} \sqrt{48\frac{\log (t/\delta)}{n_k(\tpi)}}-H^{\widetilde \pi}\frac{K(\widetilde\pi)}{n_k(\tpi)},
\end{align*}
with probability $1-(\delta/t)^6$. Inequality $(1)$ is enforced by the episode stopping condition on Line12 and the definition of $B(\pi)$, $(2)$ is guaranteed by Lem.~\ref{lem:optimism}, $(3)$ relies on the definition of gap and Assumption~\ref{asm:gap}, while $(4)$ is a direct application of Lem.~\ref{lem:conc:Markov}. Rearranging the terms, and applying Lem.~\ref{lem:episodes}, we obtain
\begin{equation*}
n_k(\tpi) \Gamma_{\min}\leq (3\widehat{H}+3+H^{\widetilde \pi}) \sqrt{n(\widetilde\pi)} \sqrt{48\log(t/\delta)}+4H^{\widetilde \pi}(2\log_2(t)+\log_2( f^{-1}(H^{+}) )).
\end{equation*}
By solving the inequality w.r.t. $n_k(\tpi)$ we obtain
\begin{equation}
\label{eq:n.bound}
\sqrt{n(\widetilde\pi)}\leq \frac{(3\widehat H+3+ H^{\widetilde \pi} )\sqrt{48\log(t/\delta)}+2\sqrt{H^{\widetilde \pi}\Gamma_{\min}(2\log_2(t)+\log_2( f^{-1}(H^{+}))}}{\Gamma_{\min}},
\end{equation}
w.p. $1-(\delta/t)^6$. This implies that on the event $\mathcal E$, after $t$ steps, RLPA acted according to a suboptimal policy $\pi$ for no more than $O(\log(t)/\Gamma_{\min}^2)$ steps. The rest of the proof follows similar steps as in Thm.~\ref{thm:gap.independent} to bound the regret of all the suboptimal policies in high probability. The expected regret of $\pi^+$ is bounded by $H^+$ and standard arguments similar to~\cite{UCRLAuer} are used to move from high-probability to expectation bounds. \qed
\end{proof}
\end{small}
Note that although the bound in Thm.~\ref{thm:gap.independent} is stated in high-probability, it is easy to turn it into a bound in expectation with almost identical dependencies on the main characteristics of the problem and compare it to the bound of Thm.~\ref{thm:gap.dependent}. 
The major difference is that the bound in Eq.~\ref{eq:gap.dependent} shows a $O(\log(T)/\Gamma_{\min})$ dependency on $T$ instead of $O(\sqrt{T})$. This suggests that whenever there is a big margin between the best policy and the other policies in $\Pi$, the algorithm is able to accordingly reduce the number of times suboptimal policies are selected, thus achieving a better dependency on $T$. On the other hand, the bound also shows that whenever the policies in $\Pi$ are very similar, it might take a long time to the algorithm before finding the best policy, although the regret cannot be larger than $O(\sqrt{T})$ as shown in Thm.~\ref{thm:gap.independent}.

We also note that while Assumption~\ref{asm:gap} is needed to allow the algorithm to ``discard'' suboptimal policies with only a logarithmic number of steps, Assumption~\ref{asm:avg.reward} is more technical and can be relaxed. It is possible to instead only require that each policy $\pi\in\Pi$ has a bounded span, $H^\pi < \infty$, which is a milder condition than requiring a constant average reward over states (i.e., $\mu^\pi(s) = \mu^\pi$).
%
%
%
%
\section{Computational Complexity}\label{s:computation}
As shown in Algo.~\ref{alg:B.MDP}, RLPA runs over multiple trials and 
episodes where  policies are selected and run. The
largest computational cost in RLPA is at the
start of each episode computing the $B$-values
for all the policies currently active in $\Pi_A$ and
then selecting the most optimistic one. This is an
$O(m)$ operation.  The total number of episodes can be
 upper bounded by $2\log_2(T)+\log_2(f^{-1} (H^+))$ (see Lem.~\ref{lem:episodes}). This means the overall computational of RLPA is 
of $O(m (\log_2(T)+\log_2(f^{-1} (H^+))))$.
Note there is no explicit dependence on the size of the state
and action space. In contrast, UCRL2 has a similar
number of trials, but requires solving extended
value iteration to compute the optimistic MDP policy.
Extended value iteration requires $O(|S|^2 |A| \log (|S|))$
computation per iteration: if $D$ are the number of
iterations required to complete extended value iteration,
then the resulting cost would be $O(D|S|^2 |A| \log (|S|)$.
Therefore UCRL2, like many generic
RL approaches, will suffer a computational complexity
that scales quadratically with the number of states,
in contrast to RLPA, which depends linearly  on the
number of input policies and is independent of the
size of the state and action space.
%
%
%
%
%
%
%
%
%
%
%
%
%
%
\section{Experiments}
In this section we provide some preliminary empirical evidence of the benefit 
of our proposed approach. We compare our approach with two other baselines. 
As mentioned previously, 
UCRL2~\cite{UCRLAuer} is a well known algorithm for generic RL problems
that enjoys strong theoretical guarantees in terms of high probability regret bounds with the optimal rate of $O(\sqrt{T})$. Unlike our approach,
UCRL2 does not make use of any policy advice, and its regret 
scales with the number of states and actions as $O(|\calS|\sqrt{|\A|})$. To provide a more 
fair comparison, we also introduce a natural variant of UCRL2, 
Upper Confidence with Models (UCWM), which takes as input a 
set of MDP models $\mathcal{M}$ which is assumed to contain the actual model $M$.
Like UCRL2, UCWM computes confidence intervals over the 
task's model parameters, but then selects the optimistic 
policy among the optimal policies for the subset of models 
in $\mathcal{M}$ consistent with the confidence interval. 
This may result in significantly tighter upper-bound on the optimal value function compared to UCRL2, 
and may also accelerate the learning process. 
If the size of possible models shrinks to one, then UCWM will seamlessly 
transition to following the optimal policy for the identified model. 
UCWM requires as input a set of MDP models, whereas our RLPA 
approach requires only input policies.

We consider a square grid world with 
$4$ actions: up ($a_1$), down ($a_2$), right ($a_3$)  and  left ($a_4$) for every state. A \emph{good} action succeeds with the probability  $0.85$, and goes in one    of the other directions with probability 0.05 (unless that would cause it to go into a wall) and 
a \emph{bad} action stays in the same place with probability $0.85$    and goes in one of the $4$ other directions with probability $0.0375$. 
We construct four variants of this grid world  
$\mathcal M=\{M_1,M_2,M_3,M_4\}$.  
In model $1$ ($M_1$) good actions are $1$ and $4$,   
in model $2$ ($M_2$) good actions are $1$ and $2$,
in model $3$ good actions are $2$ and $3$, 
and in model $4$ good actions are $3$ and $4$.
All other actions in each MDP are bad actions. 
The reward in all MDPs is the same and is $-1$ for all 
states except for the four corners which are: 0.7 (upper left), 
0.8 (upper right), 0.9 (lower left) and 0.99 (lower right). 
UCWM receives as input the MDP models and 
RLPA receives as input the optimal policies of  $\mathcal{M}$. 

We evaluate the performances of each algorithm in terms of 
the per-step regret,  
$\hat\Delta=\Delta/T$ (see Eq.~\ref{eq:regret}). Each run is $T=100000$ 
steps and we average the performance on $100$ runs. The  agent is randomly placed at one of the states of the grid at the beginning of each round. We assume that the true MDP model is $M_4$. Notice that in this case $\pi^*\in\Pi$, thus $\mu^+=\mu^*$ and the regret compares to the optimal average reward. The identity of the true MDP is not known by the agent. For RLPA we set $f(t)=\log(t)$.\footnote{See Sec. \ref{ss:gap.independent} for the rational behind this choice.} We construct grid worlds of various 
sizes and compare the resulting performance of the three algorithms. 

\begin{figure}[t!]
\begin{center}
\includegraphics[width=0.8\textwidth]{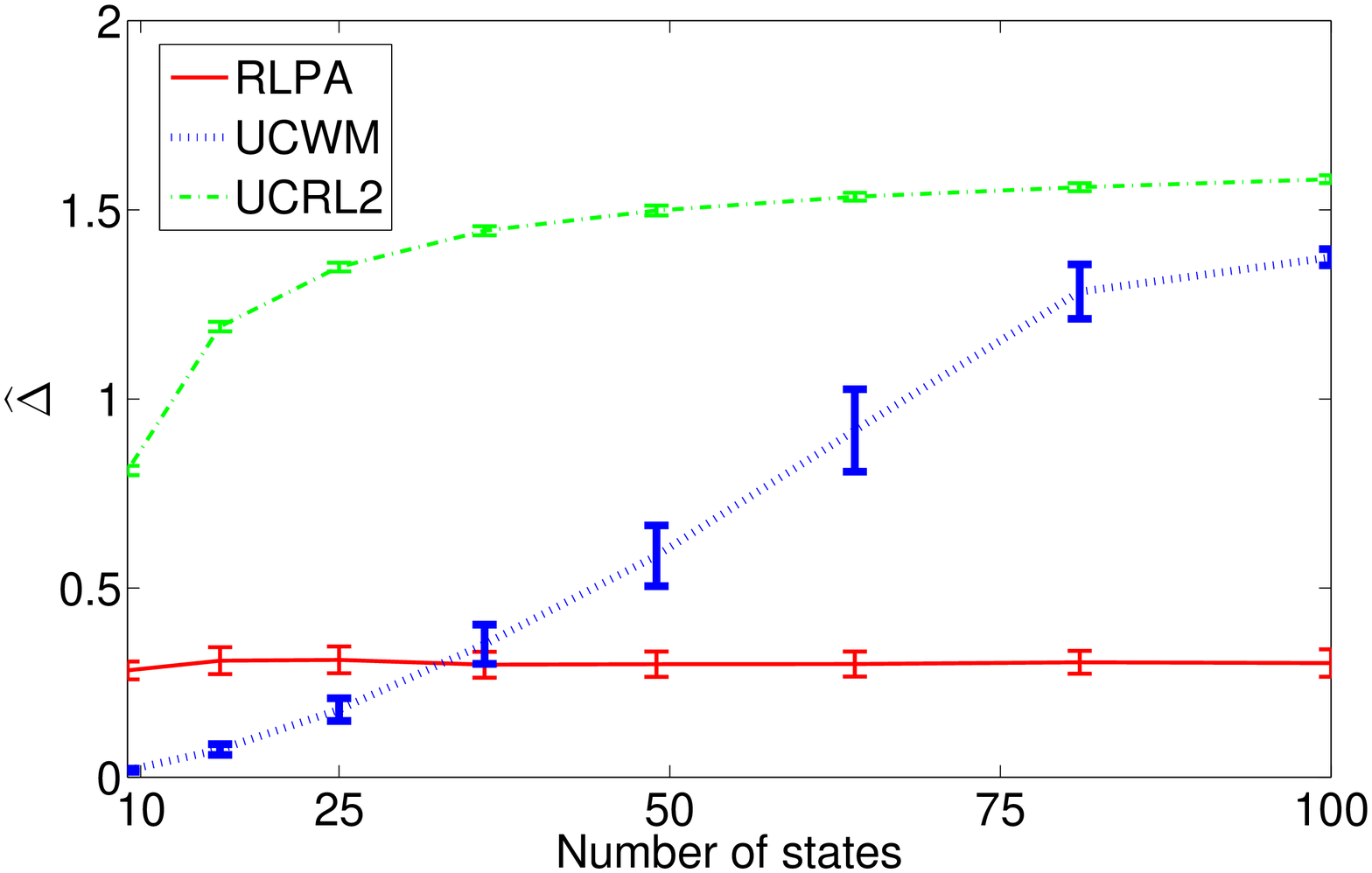}
\end{center}
\vspace{-.3in}
\caption{Per-step regret versus number of states.}
\vspace{-0.1in}
\label{fig:grid.perstate}
\end{figure}
\begin{figure}[t]
\begin{centering}
\subfigure[Avg. per-step regret vs time step.]{\label{fig:grid.perstep}
\includegraphics[width=0.5\textwidth]{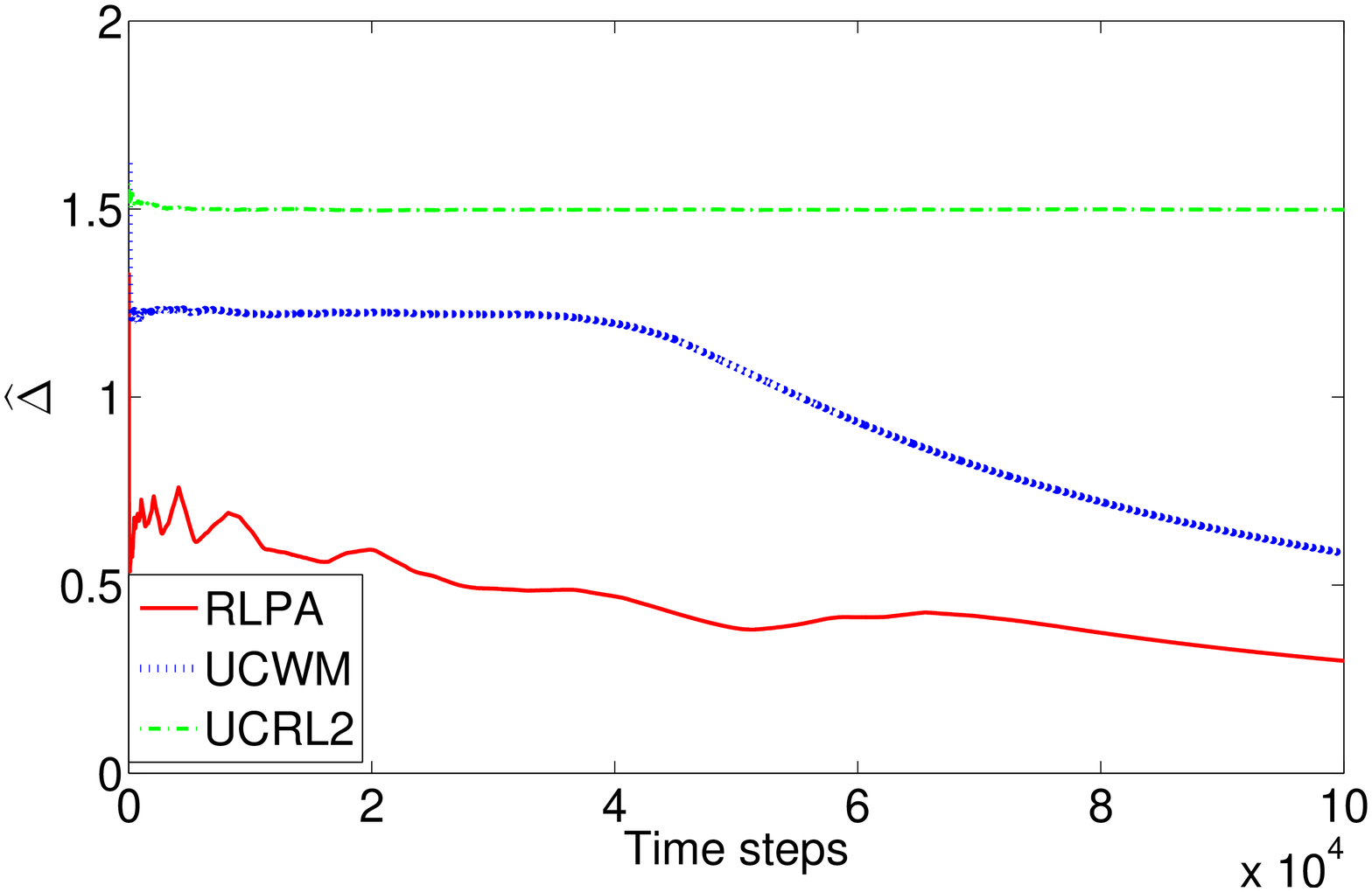}}
\subfigure[Running time versus $|S|$.]{\label{fig:grid.comp}
\includegraphics[width=0.5\textwidth]{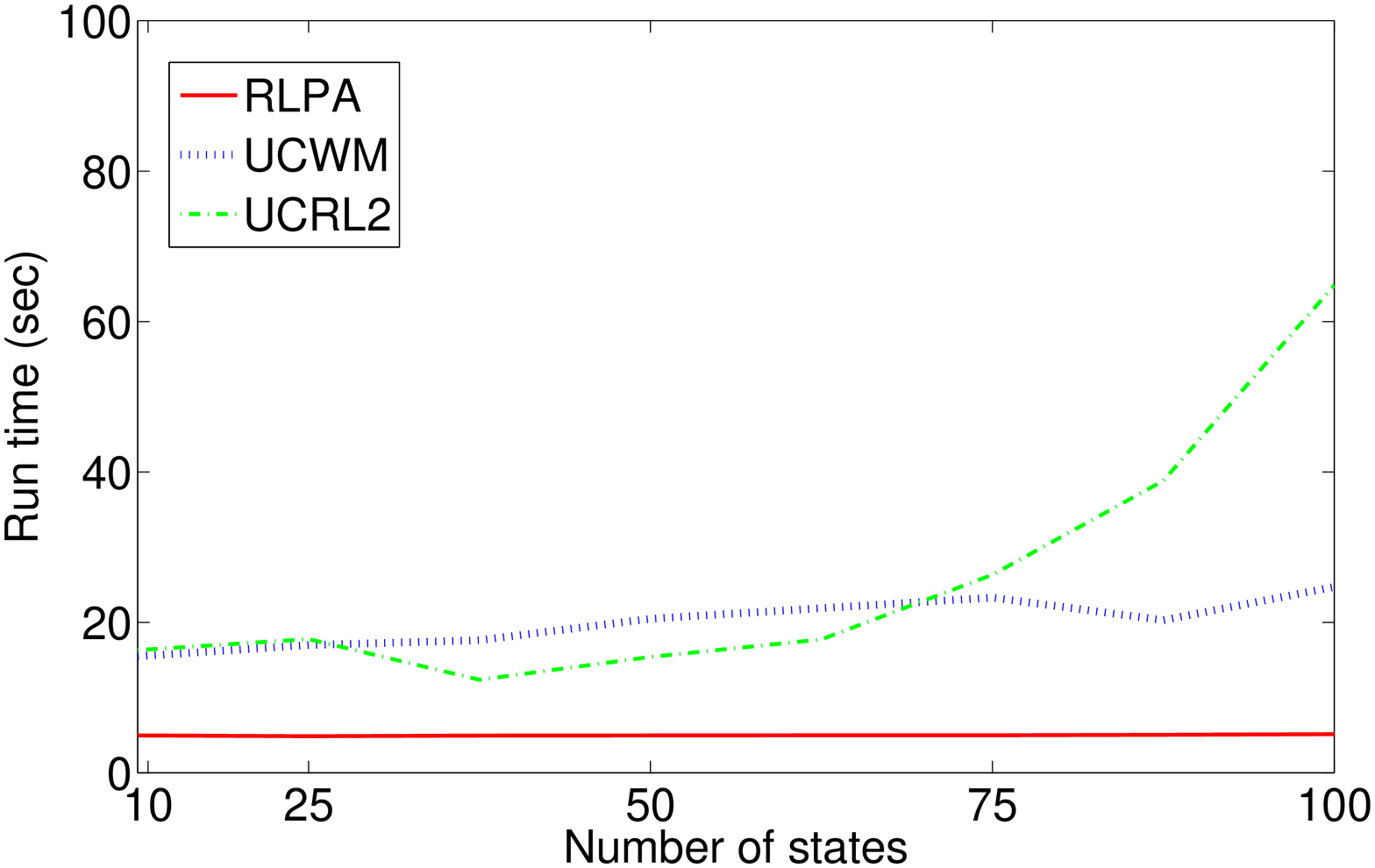}}
\end{centering}
\vspace{-0.25in}
\end{figure}



Fig.~\ref{fig:grid.perstate} shows per-step regret of the algorithms as the function of the number of states. 
As predicted by the theoretical bounds, the per-step 
regret  $\widehat\Delta$ of UCRL2 
significantly increases  as the number of states increases, 
whereas the average regret of our RLPA is essentially 
independent of the state space size\footnote{
The RLPA regret bounds depend on the bias of the 
optimal policy which may be indirectly a function 
of the structure and size of the domain.}. 
Although UCWM has a lower regret than RLPA 
for a small number of states, it quickly loses its advantage 
as the number of states grows. UCRL2's per-step regret 
plateaus after a small number of states since it is 
effectively reaching the maximum possible regret given 
the available time horizon. 

To demonstrate the performance of each approach for a 
single task, Fig.~\ref{fig:grid.perstep} shows 
how the per-step regret changes with different time horizons for a grid-world 
with $64$ states.  
RLPA demonstrates a superior regret throughout the run with a decrease 
that is faster than both UCRL and UCWM. 
The slight periodic increases in regret of RLPA are when a new trial is 
started, and all policies are again considered. 
We also note that the slow rate of decrease for all three algorithms 
is due to confidence intervals dimensioned according to the theoretical results which are often over-conservative, since 
they are designed to hold in the worst-case scenarios.
Finally, Fig.~\ref{fig:grid.comp} shows the average running time of one trial of the algorithm as a function of the number of states. As 
expected, RLPA's running time is independent of the 
size of the state space, 
whereas the running time of the other algorithms increases. 

Though a simple domain, these empirical results support our earlier 
analysis, demonstrating RLPA exhibits a regret and computational 
performance that is essentially independent of the size of the 
domain state space. This is a significant advantage over UCRL2, as we might expect because RLPA can efficiently leverage input policy advice. Interestingly, we obtain a significant improvement also over the more competitive baseline UCWM. 

\section{Related Work}
%
%


The setting we consider relates to the multi-armed bandit literature, 
where an agent seeks to optimize its reward by uncovering the arm 
with the best expected reward. More specifically, our setting 
relates to restless~\cite{ortnerALT2012} and rested~\cite{tekinIEEE2012} 
bandits, where each arm's distribution is generated by a an 
(unknown) Markov chain that either transitions at every step, 
or only when the arm is pulled, respectively. Unlike 
either restless or rested bandits, in our case each ``arm'' is 
itself a MDP policy, where different actions may be chosen. 
However, the most significant distinction may be that in our 
setting there is a independent state that couples the rewards 
obtained across the policies (the selected action depends on 
both the policy/arm selected, and the state), 
in contrast to the rested and restless 
bandits where the Markov chains of each arm evolve independently. 

Prior research has demonstrated a significant improvement 
in learning in a discrete state and action 
RL task whose Markov decision process 
model parameters are constrained to lie in a finite set. 
In this case, an objective of maximizing the expected 
sum of rewards can be framed as planning in a finite-state 
partially observable Markov decision process~\cite{PoupartICML2006}: if 
the parameter set is not too large, off-the-shelf POMDP 
planners can be used to yield 
significant performance improvements over state-of-the-art 
RL approaches~\cite{BrunskillAAMAS2012}. Other work~\cite{Dyagilev2008EWRL}
on this setting has proved that the sample complexity of 
learning to act well scales independently of the size 
of the state and action space, and linearly with the 
size of the parameter set. These approaches focus on 
leveraging information about the model space in the 
context of Bayesian RL or PAC-style RL, in contrast 
to our model-free approach that focuses on regret. 

There also exists a wealth of literature on learning 
with expert advice (e.g.~\cite{CesaACM1997}). The majority 
of this work lies in supervised learning. Prior work 
by Diuk et al.~\cite{DiukICML2009} leverages a set 
of experts where each expert predicts a probabilistic 
concept (such as a state transition) to provide 
particularly efficient KWIK RL. In contrast, our 
approach leverages input policies, rather than models. 
Probabilistic policy reuse~\cite{FernandezAAMAS2006} 
also adaptively selects among a prior set of provided policies, 
but may also choose to create and follow a new policy. 
The authors present promising empirical results but
no theoretical guarantees are provided. However,  
we will further discuss this interesting issue 
more in the future work section. 

The most closely related work is by Talvitie and 
Singh~\cite{talvitieIJCAI2007}, who also consider identifying 
the best policy from a set of input provided policies. 
Talvitie and Singh's approach is a special case of a 
more general framework for leveraging experts in 
sequential decision making environments where the 
outcomes can depend on the full history of states 
and actions~\cite{pucciNIPS2004}: however, this more 
general setting provides bounds in terms of an 
abstract quantity, whereas Talvitie and Singh provide 
bounds in terms of the bounds on mixing times of a MDP. 
There are several similarities between our 
algorithm and the work of Talvitie and Singh, 
though in contrast to their approach 
 we take an optimism under uncertainty approach, leveraging 
confidence bounds over the potential average reward of each 
policy in the current task. However, the provided bound in their 
paper is not a regret bound and no precise expression 
on the bound is stated, rendering it infeasible to do 
a careful comparison of the theoretical bounds. In contrast, 
we provide a much more rigorous theoretical analysis, 
and do so for a more general setting (for example, 
our results do not require the MDP to be ergodic). 
Their algorithm also involves several parameters whose values 
must be correctly set for the bounds to hold, but precise 
expressions for these parameters were not provided, making 
it hard to perform an empirical comparison.

\section{Future Work and Conclusion}
\label{s:conclusions}

In defining RLPA we preferred to provide a simple 
 algorithm which allowed us to provide a rigorous theoretical analysis. 
Nonetheless, we expect the current version of the algorithm 
can be easily improved over multiple dimensions. The immediate 
possibility is to perform off-policy learning across 
the policies: whenever a reward information is received for a particular 
state and action, this could be used to 
update the average reward estimate $\hmu(\pi)$ for all policies 
that would have suggested the same action for the given state.
As it has been shown in other scenarios, we expect this could 
improve the empirical performance of RLPA. However, 
the implications for the theoretical results are less clear. 
Indeed, updating the estimate $\hmu(\pi)$ of a policy $\pi$ 
whenever a ``compatible'' reward is observed would correspond to a significant increase in the number of episodes $K(\pi)$ (see Eq.~\ref{eq:empirical.average}). As a result, the convergence rate of $\hmu(\pi)$ might get worse and 
could potentially degrade up to the point when $\hmu(\pi)$ does 
not even converge to the actual average reward $\mu^\pi$. (see Lem.~\ref{lem:conc:Markov} when $K(\pi) \simeq n(\pi)$). 
We intend to further 
investigate this in the future. 


Another very interesting direction of future work is to 
extend RLPA to leverage policy advice when useful, but 
still maintain generic RL guarantees if the input policy 
space is a poor fit to the current problem. More concretely,  currently 
if 
$\pi^+$ is not the actual optimal policy of the MDP, RLPA suffers an additional linear regret to the optimal policy of order $T(\mu^*-\mu^+)$. 
If $T$ is very large and $\pi^+$ is highly suboptimal, the total regret of RLPA may be worse than UCRL, which always eventually learns the optimal policy. This opens the question whether it is possible to design an algorithm able to take advantage of the small regret-to-best of RLPA when $T$ is small and $\pi^+$ is nearly optimal and the guarantees of UCRL for the regret-to-optimal.

To conclude, we have presented RLPA, a new RL algorithm 
that leverages an input set of policies. We prove the regret 
of RLPA relative to the best policy scales sub-linearly with 
the time horizon, and that both this regret and the 
computational complexity of RLPA are independent of 
the size of the state and action space. This suggests that 
RLPA may offer significant advantages in large domains 
where some prior \emph{good} policies  are available.

\bibliography{refs.bib}
\bibliographystyle{plain}

\end{document}